\documentclass[nohyperref]{article}

\usepackage{microtype}
\usepackage{graphicx}
\usepackage{booktabs} 

\usepackage{multirow}
\usepackage{float}
\usepackage{subcaption}
\usepackage{array}
\usepackage{soul}
\newcolumntype{P}[1]{>{\centering\arraybackslash}p{#1}}
\newcolumntype{M}[1]{>{\centering\arraybackslash}m{#1}}

\usepackage[table]{xcolor}
\definecolor{lightblue}{rgb}{0.68, 0.85, 0.9}

\definecolor{pastelred}{rgb}{1.0, 0.41, 0.38}
\newcommand{\ccred}{\cellcolor{pastelred}}
\definecolor{pastelpurple}{rgb}{0.7, 0.62, 0.71}
\newcommand{\ccpurple}{\cellcolor{pastelpurple}}

\usepackage[toc,page,header]{appendix}
\usepackage{minitoc}

\usepackage{ifthen}
\usepackage[normalem]{ulem}
\newcommand{\niv}[2][]{%
    \ifthenelse{ \equal{#1}{} }
        {\textcolor{red}{(NH) #2}}
        {\textcolor{red}{(NH) \sout{#1} #2}}
}
\newcommand{\yaniv}[2][]{%
    \ifthenelse{ \equal{#1}{} }
        {\textcolor{green}{(YN) #2}}
        {\textcolor{green}{(YN) \sout{#1} #2}}
}

\usepackage{hyperref}

\usepackage[accepted]{icml2023}

\usepackage{amsmath}
\usepackage{amssymb}
\usepackage{mathtools}
\usepackage{amsthm}

\usepackage[capitalize,noabbrev]{cleveref}

\theoremstyle{plain}

\theoremstyle{definition}

\theoremstyle{remark}

\usepackage[textsize=tiny]{todonotes}

\icmltitlerunning{SinFusion: Training Diffusion Models on a Single Image or Video}

\begin{document}

\twocolumn[
\icmltitle{SinFusion: Training Diffusion Models on a Single Image or Video}

\icmlsetsymbol{equal}{*}

\begin{icmlauthorlist}
\icmlauthor{Yaniv Nikankin}{wis,equal}
\icmlauthor{Niv Haim}{wis,equal}
\icmlauthor{Michal Irani}{wis}
\\\icmlauthor{\raisebox{-.2cm}{Project Page:~~\href{https://yanivnik.github.io/sinfusion}{\color{blue}{https://yanivnik.github.io/sinfusion}}}}{}
\end{icmlauthorlist}

\icmlaffiliation{wis}{Department of Computer Science and Applied Mathematics,
Weizmann Institute of Science, Rehovot, Israel}

\icmlcorrespondingauthor{Yaniv Nikankin}{yaniv.nikankin@weizmann.ac.il}

\icmlkeywords{Single Video Generation, Generative Models, Machine Learning, ICML}

\vskip 0.3in
]

\printAffiliationsAndNotice{\icmlEqualContribution}

\begin{figure*}[!ht]
\vspace*{-0.3cm}
    \centering
    \includegraphics[width=\textwidth]{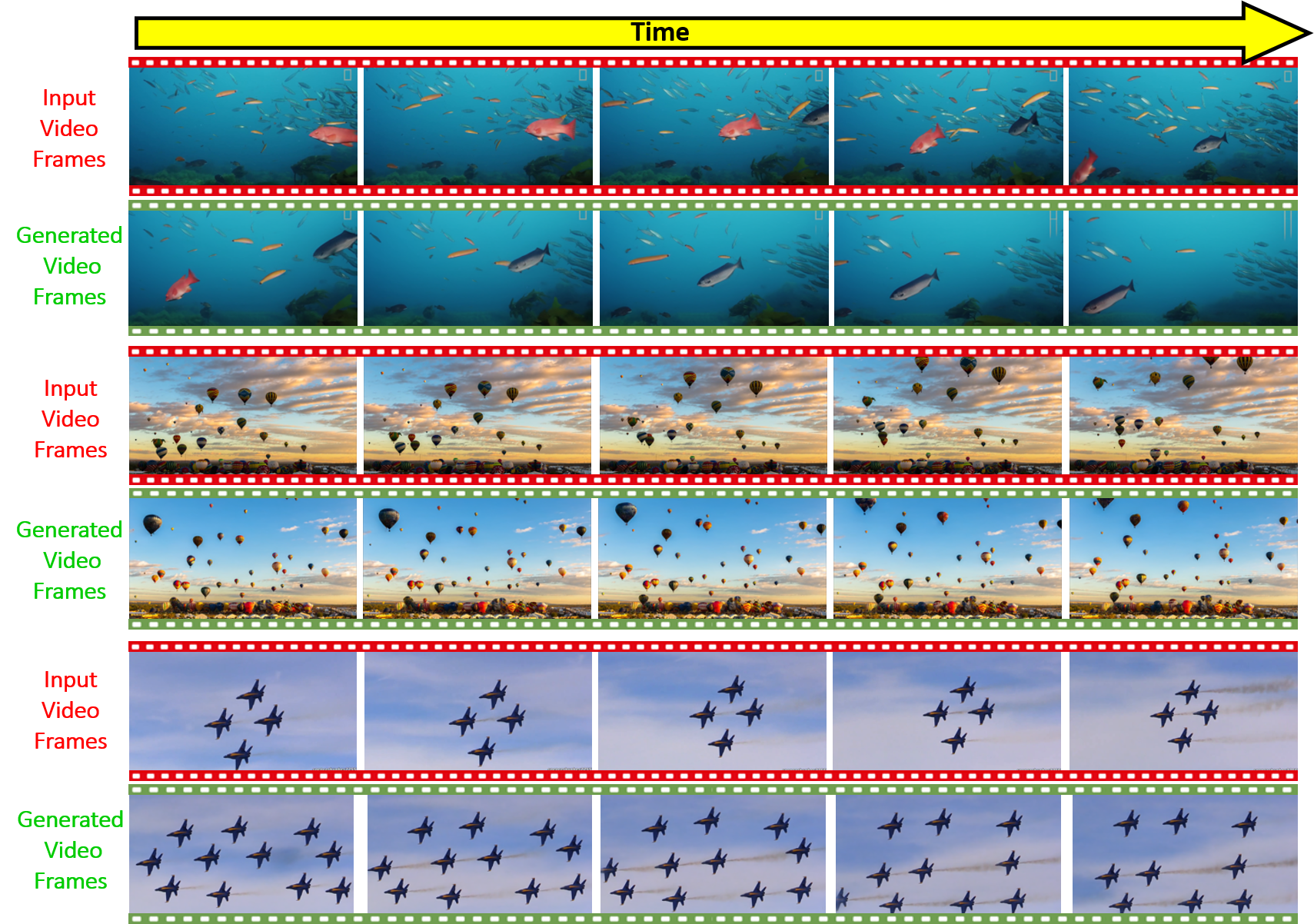}
 \vspace*{-0.7cm}
   \caption{\textbf{Diverse video generation.} For each single training video, \textit{red row} shows consecutive frames from the training video, whereas the \textit{green row} show a set of consecutive frames generated by our single video DDPM. 
    \textbf{Please see the videos in our \href{https://yanivnik.github.io/sinfusion/}{project page}}.
    \vspace{-8pt}
    }
    \label{fig:video_results_diverse_generation}
\end{figure*}

\begin{abstract}
Diffusion models exhibited tremendous progress in image and video generation, exceeding GANs in quality and diversity. However, they are usually trained on very large datasets and are not naturally adapted to manipulate a given input image or video. In this paper we show how this can be resolved by training a diffusion model on a single input image or video. Our image/video-specific diffusion model (SinFusion) learns the appearance and dynamics of the single image or video, while utilizing the conditioning capabilities of diffusion models. It can solve a wide array of image/video-specific manipulation tasks. In particular, our model can learn from few frames the motion and dynamics of a single input video. It can then generate diverse new video samples of the same dynamic scene, extrapolate short videos into long ones (both forward and backward in time) and perform video upsampling. Most of these tasks are not realizable by current video-specific generation methods.

\end{abstract}

\section{Introduction}
\label{sec:intro}

Until recently, generative adversarial networks (GANs)
ruled the field of generative models, with seminal works like StyleGAN~\cite{karras2017progressive,karras2019style,karras2020analyzing}, BigGAN~\cite{brock2018large} etc.~\cite{radford2015unsupervised,zhang2019self}. 
Diffusion models (\textbf{DMs})~\cite{sohl2015deep,song2019generative,ho2020denoising} have gained the lead in the last years, surpassing GANs by image quality and diversity~\cite{dhariwal2021diffusion} and becoming the leading method in many vision tasks like text-to-image generation, superresolution and many more~\cite{jolicoeur2020adversarial,nichol2021improved,song2020denoising,saharia2022image,ho2022cascaded,nichol2021glide,saharia2022photorealistic,rombach2022high} (see surveys~\cite{cao2022survey,croitoru2022diffusion}). Recent works also demonstrate the effectiveness of DMs for video and text-to-video generation~\cite{ho2022video,singer2022make,ho2022imagen,villegas2022phenaki}.

DMs are trained on massive datasets and as such, these models are very large and resource demanding.
Applying their capabilities to edit or manipulate a specific input provided by the user is non-trivial and requires careful manipulation and fine-tuning 
\cite{avrahami2022blended,gal2022textual,ruiz2022dreambooth,valevski2022unitune,kawar2022imagic}.

In this work we propose a framework for training diffusion models on a single input image or video - \emph{``SinFusion''}. We harness the success and high-quality of DMs at image synthesis, to single-image/video tasks. 
Once trained, SinFusion can generate new image/video samples with similar appearance and dynamics to the original input and perform various editing and manipulation tasks. In the video case, SinFusion exhibits impressive generalization capabilities by coherently extrapolating an input video far into the future (or past). 
This is learned from very few frames (mostly 2-3 dozens, but is already apparent for fewer frames).

We demonstrate the applicability of SinFusion to a variety of single-video tasks, including: (i)~diverse generation of new videos from a \emph{single} input video (better than existing methods), (ii)~video extrapolation (both forward and backward in time), (iii)~video upsampling. Many of these taks (e.g., extrapolation/interpolation in time) are not realizable by current video-specific generation methods~\cite{gur2020hierarchical,haim2021diverse}. Moreover, large-scale diffusion models for video generation~\cite{yang2022diffusion,ho2022video} trained on large video datasets are not designed to manipulate a  real input video.
When applied to a single input image, SinFusion can perform diverse image generation
and manipulation tasks.
However, the main focus in our paper is on \emph{single-video}  generation/manipulation tasks, as this is a more challenging and less explored domain.

Our framework builds on top of the commonly used DDPM architecture~\cite{ho2020denoising}, but introduces several important modifications that are essential for allowing it to train on a single image/video. Our backbone DDPM network is \emph{fully convolutional}, hence can be used to generate images of any size by starting from a noisy image of the desired output size. Our single-\emph{video} DDPM, consists of $3$ single-\emph{image} DDPMs, each trained to map noise to large crops of an image (a video frame), either unconditionally, or conditioned on other frames from the input video.

\textbf{Our main contributions are as follows:} \\
$\bullet$ First-ever diffusion model trained on a single image/video. \\
$\bullet$  Unlike general large-scale diffusion models, SinFusion can edit and manipulate a \emph{real input video}. This includes: diverse video generation, video extrapolation (both forward and backward in time), and temporal upsampling. \\
$\bullet$   SinFusion provides new video capabilities and tasks  not realizable by current single-video GANs (e.g., video extrapolation with impressive motion generalization capabilities). \\
$\bullet$  We propose a new set of evaluation metrics for diverse video generation from a single video.

\section{Related Work}

Our work lies in the intersection of several fields: generative models trained on a single image or video, manipulation of a real input image/video, diffusion models and methods for image/video generation in general. Here we briefly mention the main achievements in each field and their relation (and difference) from our proposed approach.

\textbf{Video generation} is a broad field of research
including many areas such as video GANs~\cite{vondrick2016generating, tulyakov2018mocogan,clark2019adversarial,skorokhodov2022stylegan}, video-to-video translation~\cite{wang2018video,bansal2018recycle} or autoregressive prediction models~\cite{ballas2015delving,villegas2017decomposing, babaeizadeh2017stochastic, denton2018stochastic}, to name a few. Diffusion models for video generation are fairly recent and mostly rely on DDPM~\cite{ho2020denoising} framework for image generation, extended to handle videos \cite{yang2022diffusion,hoppe2022diffusion,voleti2022mcvd,ho2022video,ho2022imagen,ho2022imagen,harvey2022flexible} (see Appendix~\ref{sec:diffusion_videos}).
These methods can synthesize beautiful videos,
however, none of them can modify or manipulate an existing input video provided by the user, which is our goal. 

\begin{figure*}[!htb]
\vspace*{-0.3cm}
    \centering
    \includegraphics[width=.9\textwidth]{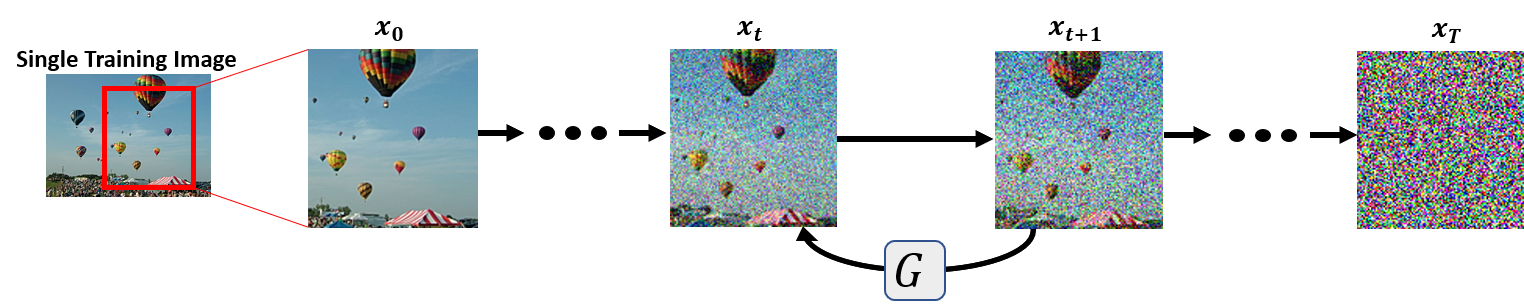}
\vspace*{-0.5cm}
    \caption{\textbf{Single Image DDPM.} Our single-image DDPM trains on large crops from a single image. It learns to remove noise from noisy crops, and, at inference, can generate diverse samples with similar structure and appearance to the training image.
    }
    \label{fig:image_ddpm_overview}
    \centering
\vspace*{-0.3cm}
\end{figure*}

\textbf{Generative Models trained on a Single Image or Video} aim to generate new diverse samples, similar in appearance and dynamics to the image/video on which they were trained. Most notably, SinGAN~\cite{shaham2019singan} and InGAN~\cite{shocher2018ingan} trained multi-scale GANs to learn the distribution of patches in an image. They showed its applicability to diverse random generation from a single image, as well as a variety of other image synthesis applications (inpainting, style transfer, etc.). However, GPNN~\cite{granot2021drop} showed that most image synthesis tasks proposed by single-image GAN-based models can be solved by classical non-parametric patch nearest-neighbour methods~\cite{efros1999texture,efros2001image,simakov2008summarizing}, and achieve outputs of higher quality while reducing generation time by orders of magnitude. Similarly, extensions of SinGAN~\cite{shaham2019singan} to generation from a single \emph{video}~\cite{gur2020hierarchical,arora2021singan} were outperformed by patch nearest-neighbour methods~\cite{haim2021diverse}.
However, nearest-neighbour methods have a very limited notion of generalization and are therefore limited to tasks where it is natural to ``copy" parts of the input.
While  generated samples are of high quality and look realistic, this is because the samples are essentially \emph{copies} of parts of the original video stitched together. \emph{They fail to exhibit motion generalization capabilities}. 
In contrast, our method generalizes well from just a few frames and can be easily trained on a long input video.
Concurrently to our work,~\citet{kulikov2022sinddm,wang2022sindiffusion} trained DMs on a single image and showed various capabilities. However, both works focused on generation from a single \emph{image}, while we present applications on a single \emph{video}.

\vspace*{-0.3cm}
\paragraph{Reference Image Manipulation with Large Generative Models.}
One of the practical application of generative models trained on large datasets is their strong generalization capabilities for semantic image editing, often obtained via latent space interpolation~\cite{radford2015unsupervised,brock2018large,karras2019style}. Applying these capabilities to an existing reference image was mostly achieved by GAN “inversion” techniques~\cite{xia2022gan}, and very recently by fine-tuning large diffusion models \cite{gal2022textual,kawar2022imagic,ruiz2022dreambooth,valevski2022unitune,avrahami2022blended}. 
However, to the best of our knowledge, there are no existing large-scale models to-date which can manipulate an existing input reference video.

\section{Perliminaries: Overview of DDPM}
\label{sec:preliminaries}
Denoising diffusion probabilistic models (DDPM) \cite{ho2020denoising,sohl2015deep}
are a class of generative models that can learn to convert unstructured noise to samples from a given distribution, by performing an iterative process of removing small amounts of Gaussian noise at each step.
Since our method heavily relies on DDPM, we provide here a very brief overview of DDPM and its basics.\\
To train a DDPM, an input image $x_0$ is sampled, and small portions of gaussian noise $\epsilon$ are gradually added to it in a parameter-free \textit{forward} process, resulting in a noisy image $x_t$. 
The forward process can be written as:
\begin{equation}
    \mathbf{x}_t=\sqrt{\bar{\alpha}_t} \mathbf{x}_0+\sqrt{1-\bar{\alpha}_t} \epsilon
\end{equation}
where $\bar\alpha_t=\prod_{s=1}^t(1-\beta_s)$, $\beta_t \in (0,1)$ is a predefined parameter and $\epsilon \sim \mathcal{N}(\mathbf{0}, \mathbf{I})$ is the noise used to generate the noisy image $x_t$.\\
A neural network is then trained to perform the \textit{reverse} process. In the reverse process, the noisy image $x_t$ is given as input to the neural network, which predicts the noise $\epsilon$ that was used to generate the noisy image. The network is trained with an $L_2$ loss:
\begin{equation}
\label{orig_ddpm_loss}
    L(\theta) = \mathbb{E}_{\mathbf{x}_0, \epsilon}\left[\left\|\epsilon-\epsilon_\theta\left(x_t, t\right)\right\|^2\right] \ .
\end{equation}
In existing DDPM-based methods, The network is typically trained on a large dataset of images, from which $x_0$ is sampled. Once trained, the generation process is initiated with a random noise image $x_T \sim \mathcal{N}(\mathbf{0}, \mathbf{I})$. The image is passed through the model in a series of \textit{reverse} steps. In each timestep $t=T,...,1$, the neural network predicts the noise $\epsilon_t$. This noise is then used to generate a less noisy version of the image ($x_{t-1}$), and the process is repeated until a possible clean image $x_0$ is generated.

\section{Single Image DDPM}
\label{sec:single_image_ddpm}

Our goal is to leverage the powerful mechanism of diffusion models to generation from a single image/video. While the main contribution of this paper is in using DDPMs for generation from a single \emph{video}, we first explain how a diffusion model can be trained on a single \emph{image}. In Sec.~\ref{sec:single_video_ddpm} we show how this model can be extended to \emph{video} generation. 
Some applications of single image DDPM are found in Sec.~\ref{sec:image_applications}.

Given a single input image, we want our model to generate new diverse samples that are similar in appearance and structure to that of the input image, but also allow for semantically coherent variations. 
We build upon the common DDPM~\cite{ho2020denoising} framework (\cref{sec:preliminaries}) and introduce several modifications to the training procedure and to the core network of DDPM. These are highlighted below:

\vspace*{-0.3cm}
\paragraph{Training on Large Crops.} Instead of training on a large collection of images, we train a single diffusion model on many large random crops from the input image (typically, about $95$\% the size of the original image, \cref{fig:image_ddpm_overview}). We find that training on the original resolution of the image is sufficient for generating diverse image samples, even without the use of multi-scale pyramid (unlike most previous single image/video generative methods \cite{arora2021singan,shocher2018ingan,shaham2019singan,hinz2021improved,gur2020hierarchical,granot2021drop,haim2021diverse}). By training on large crops our generated outputs retain the global structure of the input image.

\begin{figure}
    \centering
    \includegraphics[width=.7\columnwidth]{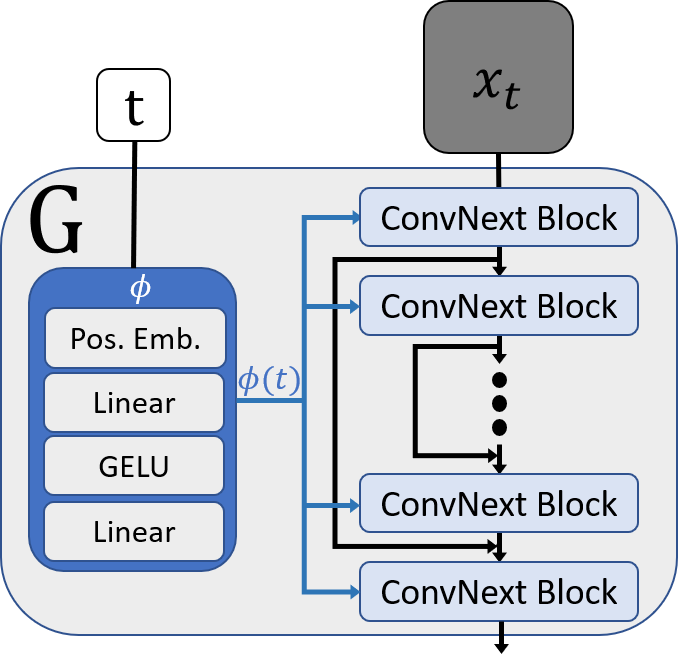}
\vspace*{-0.5cm}
    \caption{\textbf{Network Architecture}. Our backbone network is a fully convolutional chain of ConvNext~\cite{liu2022convnet} blocks with residual connections. Note that our network does not include any reduction in the spatial dimensions along the layers.}
    \label{fig:backbone}
\vspace*{-0.3cm}
\end{figure}

\vspace*{-0.3cm}
\paragraph{Network Architecture.}
\label{sec:backbone}
Directly training the standard DDPM~\cite{ho2020denoising}
on the single image or its large crops results in "overfitting", namely the model only generates the same image crops.
We postulate that this phenomenon occurs because of the receptive field of the core backbone network in DDPM, which is the entire input image. 
To this end we modify the backbone UNet~\cite{ronneberger2015u} network of DDPM, in order to reduce the size of its receptive field. 
We remove
the attention layers as they have global receptive field. We also remove the downsampling and upsampling layers which cause the receptive field to grow too rapidly. 
Removing the attention layers has an unwanted side-effect - harming the performance of the diffusion model. 
\citet{liu2022convnet} proposed a fully convolutional network that matches the attention mechanism on many vision tasks. Inspired by this idea, we replace the ResNet~\cite{he2016deep} blocks in the network with ConvNext~\cite{liu2022convnet} blocks. This architectural choice is meant to replace the functionality of the attention layers, while keeping a non-global receptive field. It also has the advantage of reducing computation time. The overall receptive field of our network is then determined by the number of ConvNext blocks in the network. Changing the number of ConvNext blocks allows us to control the diversity of the output samples. 
Please see further analysis and hyperparameter choice in \cref{sec:effect_of_cs_and_rf}.
The rest of our backbone network is similar to DDPM, as well as the embedding network ($\phi$) which is used to incorporate the diffusion timestep $t$ into the model (and will be later used to embed the video frame difference, see~\cref{sec:predictor}). See \cref{fig:backbone} for details.

\vspace*{-0.3cm}
\paragraph{Loss.}
At each training step, the model is given a noisy image crop $x_t$. 
However, in contrast to DDPM~\cite{ho2020denoising}, whose model predicts the added noise (as in \cref{orig_ddpm_loss}), our model predicts the clean image crop $\Tilde{x}_{0,\theta}$. The loss in our single-image DDPM is:
\begin{equation}
\label{our_loss}
    L(\theta) = \mathbb{E}_{\mathbf{x}_0, \epsilon}\left[\left\|x_0 - \Tilde{x}_{0,\theta}\left(x_t, t\right)\right\|^2\right]
\end{equation}
We find that predicting the image instead of the noise leads to better results when training on a single image, both in terms of quality and training time. We attribute this difference to the simplicity of the data distribution in a single image compared to the data distribution of a large dataset of images. 
The full training algorithm is as follows:

\begin{algorithm}
    \caption{Training on a single image $x$}
    \label{alg_training}
    \begin{algorithmic}[1]
        \REPEAT
            \STATE $x_0 \gets Crop(x) $
            \STATE $t \sim $ Uniform(${1,...,T=50}$)
            \STATE $\epsilon \sim \mathcal{N}(0, \mathbf{I})$
            \STATE Take gradient descent step on: \par
            $\nabla_\theta\left\|x_0-\Tilde{x}_{0,\theta}\left(\sqrt{\bar{\alpha}_t} \mathrm{x}_0+\sqrt{1-\bar{\alpha}_t} \epsilon, t\right)\right\|^2$\par
        \UNTIL converged
    \end{algorithmic}
\end{algorithm}

Our single-image DDPM can be used for various image synthesis tasks like diverse generation (\cref{fig:SingleImageTasks}), generation from sketch and image editing.

\begin{figure*}[!htb]
    \centering
    \includegraphics[width=.8\textwidth]{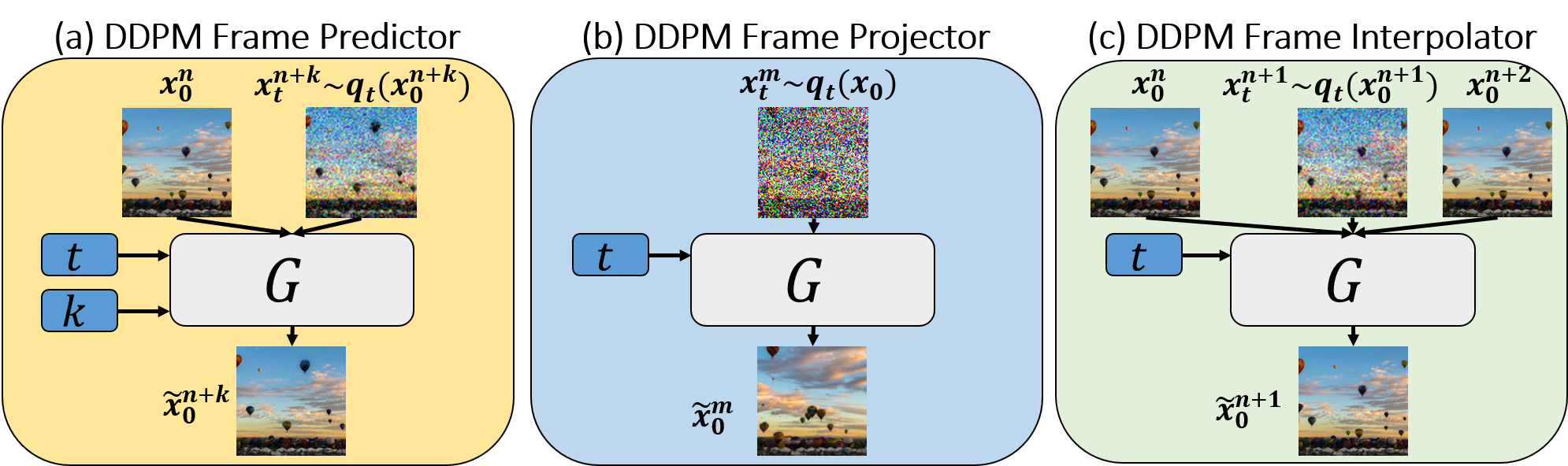}
    \vspace{-0.3cm}
    \caption{\textbf{Single Video DDPM} Our video framework consists of three models.
    The \emph{Predictor} (left) generates new frames, conditioned on previous frames.
    The \emph{Projector} (middle) generates frames from noise, and corrects small artifacts in predicted frames.
    The \emph{Interpolator} (right) interpolates between adjacent frames (conditioned on them), to upsample the video temporally.
    These models are used together at inference to perform various video related applications.
    }
    \label{fig:video_training}
    \centering
    \vspace{-0.4cm}
\end{figure*}

\section{Single Video DDPM}
\label{sec:single_video_ddpm}

Our video generation framework consists of 3 single-image-DDPM models (Fig.~\ref{fig:video_training}), whose combination gives rise to a variety of different video-related applications (Sec.~\ref{sec:applications}).
Our framework is essentially an autoregressive video generator. Namely, we train the models on a given input video with frames $\{x_0^1, x_0^2, ..., x_0^N\}$, and generate new videos with frames $\{\tilde{x}_0^1, \tilde{x}_0^2, ..., \tilde{x}_0^M\}$ such that each generated new frame $\tilde{x}_0^{n+1}$ is conditioned on its previous frame $\tilde{x}_0^{n}$.
The three models that constitute our framework are all single-image DDPM models with the same network architecture as described in Sec.~\ref{sec:single_image_ddpm}. The models are trained \emph{separately} and differ by the type of inputs they are given, and by their role in the overall generation framework. The inference is application-dependant and is discussed in Sec.~\ref{sec:applications}. Here we describe the training procedure of each model:

\vspace*{-0.25cm}
\paragraph{DDPM Frame Predictor \ (Fig.~\ref{fig:video_training}a).}
\label{sec:predictor}
The role of the \emph{Predictor} model is to generate new frames, each conditioned on its previous frame. At each training iteration we sample a condition frame from the video $x^n_0$ and a noisy version of the $(n+k)$'th frame ($x^{n+k}_t$), which is to be denoised. The two frames are concatenated along the channels axis before being passed to the model (as in~\citet{saharia2022image}).
The model is also given an embedding of the temporal difference (i.e frame index difference) between the two frames ($\phi(k)$). This embedding is concatenated to the timestep embedding ($\phi(t)$) of the DDPM. At early training $k$=$1$, and in following iterations it is gradually increased to be sampled at random from $k=[-3,3]$. We find that such a curriculum learning approach improves outputs quality (even when at inference $k$=$\pm 1$).

\vspace*{-0.25cm}
\paragraph{DDPM Frame Projector \ (Fig.~\ref{fig:video_training}b).}
\label{sec:corrector} The role of the \emph{Projector} model is to “correct” frames that were generated by the \emph{Predictor}. The Projector is a straightforward single-image-DDPM as described in~\cref{sec:single_image_ddpm}, only it is trained on image crops from \emph{all} the frames in the video.
After learning the image structure and appearance of the video frames it is used to correct small artifacts in the generated frames, that may otherwise accumulate and destroy the video generation process. Intuitively, it “projects” patches from the generated frames back unto the original patch distribution, hence its name. The Projector is also used to generate the first frame. Frame correction is done at inference via a truncated diffusion process on the predicted frame.

\vspace*{-0.2cm}
\paragraph{DDPM Frame Interpolator \ (Fig.~\ref{fig:video_training}c).}
\label{sec:interpolator}

Our video-specific DDPM framework can be further trained to increase the temporal resolution of our generated videos, known also as “video upsampling” or “frame interpolation”. Our DDPM frame \emph{Interpolator}
receives as input a pair of clean frames ($x_0^n$, $x_0^{n+2}$) as conditioning, and a noised version of the frame between them ($x_t^{n+1}$). The frames are concatenated along the channels axis, and the model is trained to predict the clean version of the interpolated frame ($\tilde{x}_0^{n+1}$). We find that this interpolation generalizes well to small motions in the video, and can be used to interpolate between every two consecutive frames, thus increasing the temporal resolution of generated videos as well as the input video.

\vspace*{-0.2cm}
\paragraph{Losses.}
We find that some models work better with different losses. The Projector and the Interpolator are trained with the loss in Eq.~(\ref{our_loss}), while the Predictor is trained with Eq.~(\ref{orig_ddpm_loss}), i.e., the noise is predicted instead of the output.

\begin{figure*}[t!]
\vspace{-0.5cm}
    \centering
    \includegraphics[width=1.05\textwidth]{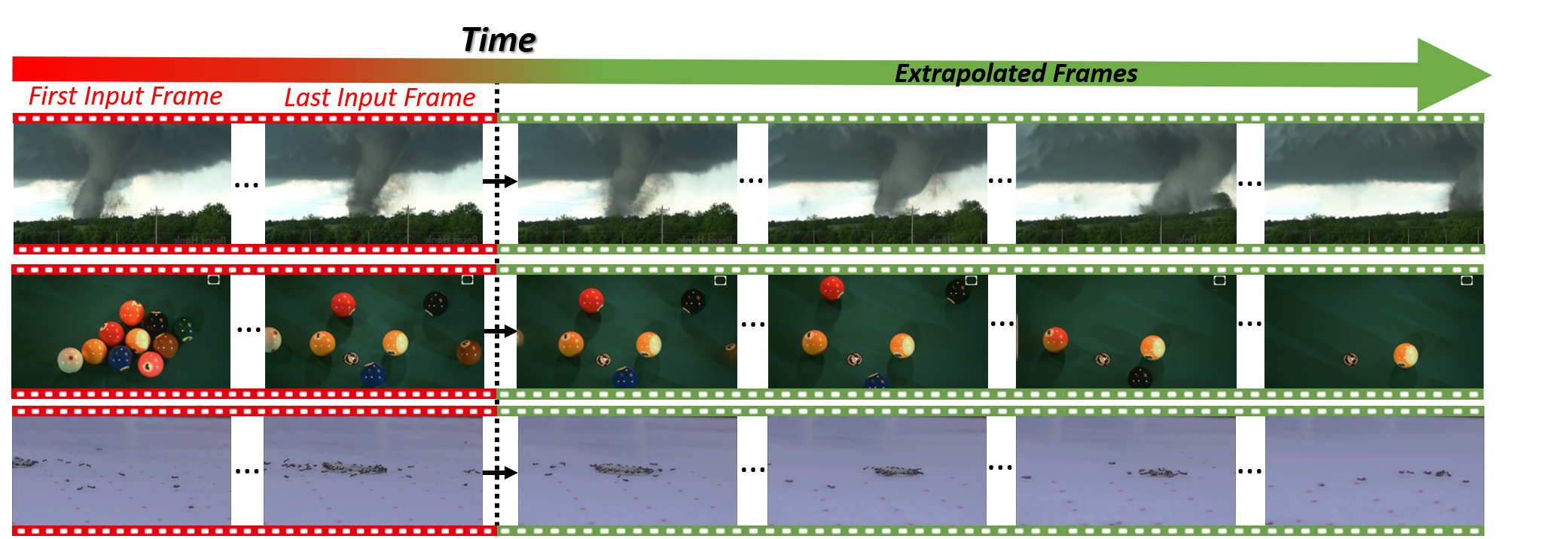}
 \vspace{-0.75cm}   
 \caption{
    \textbf{Video Extrapolation (\underline{into the Future}):} 
    \emph{SinFusion} trains on a \underline{single} input video (\textit{red})  - exemplified on video frames of Tornado, Balls, Ants. At inference, the auto-regressive generation process starts from the \emph{last} frame of the input video,
    and generates a frame sequence of any desired length. The extrapolated frames (\textit{green}) were never seen in the original video. 
    \emph{See full videos in our \href{https://yanivnik.github.io/sinfusion/}{project page.}}
    } 
    \label{fig:video_results_extrapolation}
    \centering
\vspace{-0.3cm}
\end{figure*}

\section{Applications}
\label{sec:applications}

In this section we show how combinations of our single image/video DDPMs (\cref{sec:single_image_ddpm,sec:single_video_ddpm}) provide a variety of video synthesis tasks. 
\emph{We refer the reader to our \href{https://yanivnik.github.io/sinfusion/}{project page}}, especially to view our video results.

\paragraph{Diverse Video Generation:}
We can generate diverse videos from a single input video, to any length, such that the output samples have similar appearance, structure and motions as the original input video. This is done by combining our Predictor and Projector models.
The first frame is either some frame from the original video, or a generated output image from the unconditional Projector.
The Predictor is then used to generate the next frame, conditioned on the previous generated frame. 
Next, the predicted frame is corrected by the Projector (to remove small artifacts that may have been created, thus preventing error accumulation over time). This process is repeated until the desired number of frames has been generated. Repeating this autoregressive generation process creates a new video of arbitrary length. 
Note that the process is inherently stochastic -- even if the initial frame is the same, different generated outputs will quickly diverge and create different videos.
See Fig.~\ref{fig:video_results_diverse_generation} and \emph{our \href{https://yanivnik.github.io/sinfusion/}{project page} for live videos and many more examples}.

\vspace{-0.3cm}
\paragraph{Video Extrapolation} (\emph{into the Future and into the Past}):
Given an input video, we can \emph{``predict the future''} (i.e., predict its future frames) by initializing the generation process described above with the last frame of the input video. Fig.~\ref{fig:video_results_extrapolation}  shows a few such examples.
Note how our method extrapolates the motion in a realistic way, preserving the appearance and dynamics of the original video. \textbf{To the best of our knowledge, no existing single-video generation method can extrapolate a video in time.}
Since our Predictor is also trained backward in time (predicting the previous frame using negative $k$), it can also \emph{extrapolate videos backwards in time} (\emph{``predict the past''}) by starting from the first frame of the video. This e.g. causes flying balloons to ``land'' (see video in our project page), even though these motions were never observed in the original video. This is a straightforward manifestation of the generalization capabilities of our framework. 
See Sec.~\ref{sec:evaluations} for  evaluations of the generalization capabilities, and \emph{full videos in our \href{https://yanivnik.github.io/sinfusion/}{project page}}.

\vspace{-0.3cm}
\paragraph{Temporal Upsampling:}
Not only can SinFusion \emph{extrapolate} input videos, it can also \emph{interpolate} them -- generate new frames in-between the original ones. This is done by training the DDPM Frame Interpolator (Fig.~\ref{fig:video_training}c) to predict each frame from its 2 
\emph{neighboring} frames, and  at inference applying it to interpolate between \emph{successive} frames.  The appearance of the interpolated frames is  corrected by the DDPM \mbox{Frame Projector. \emph{See example videos in our \href{https://yanivnik.github.io/sinfusion/}{project page}}.}

\begin{figure}
    \vspace*{-0.3cm}
    \centering
    \begin{tabular}{c}
        \includegraphics[width=\columnwidth]{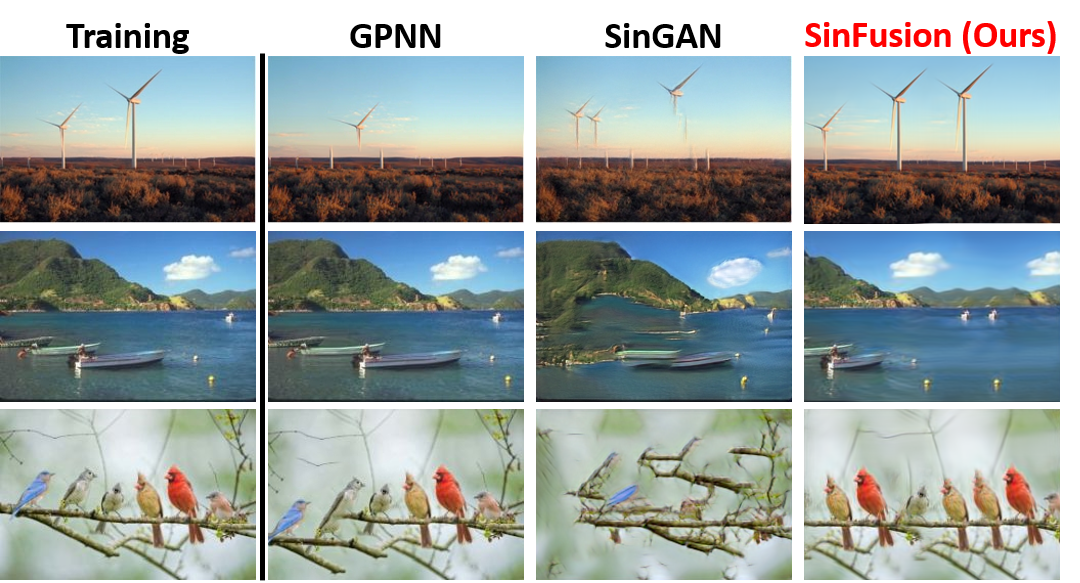} \\
        (a) Diverse generation from a single image. \\
        \includegraphics[width=\columnwidth]{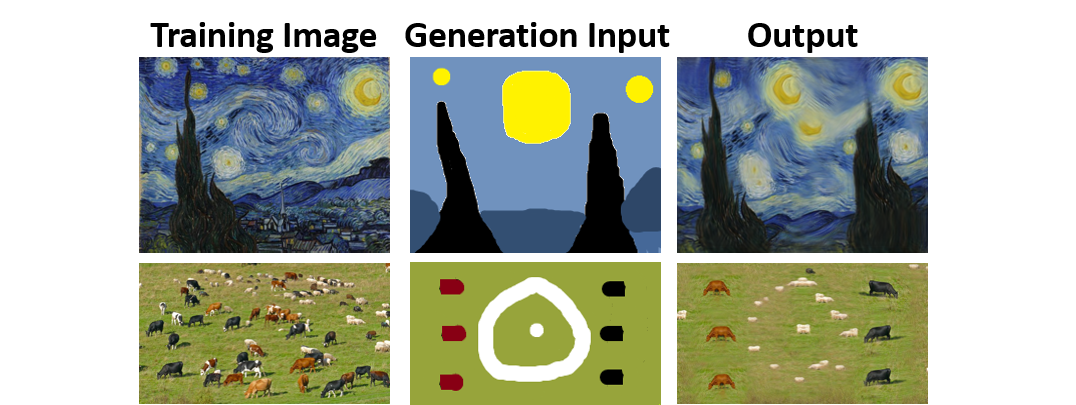} \\
        (b) Sketch-guided image generation.
    \end{tabular}
    \vspace*{-0.4cm}
    \caption{\textbf{Single Image Applications}: (a)~Images generated by SinFusion are comparable in visual quality to the patch nearest-neighbour based method GPNN~\cite{granot2021drop}, and outperforms SinGAN~\cite{shaham2019singan}.
   (b)~SinFusion can generate new images from a single image,  conditioned on input sketches.
    }
    \label{fig:SingleImageTasks}
    \vspace*{-0.4cm}
\end{figure}

\vspace{-0.3cm}
\paragraph{Single-Image Applications}
\label{sec:image_applications}
When training our single-image DDPM (Sec.~\ref{sec:single_image_ddpm}) on a single input image, our framework reduces to standard single-image generation and manipulation tasks, including: Diverse image generation, Sketch-guided image generation and Image editing. Diverse image generation is done by sampling a noisy image $x_T$$\sim$$\mathcal{N}(0,\mathbb{I})$ and iteratively denoising using our trained model such that $x_{t-1}$$=$$G(x_t)$. Since our backbone DDPM network is fully convolutional, it can be used to generate images of any size by starting from a noisy image of the desired size.
Fig.~\ref{fig:SingleImageTasks}a
shows such results (visually compared to SinGAN~\cite{shaham2019singan} and GPNN~\cite{granot2021drop}). 
\textit{See more results in our project page}.
SinFusion can also edit an input image by coarsely moving crops between locations in the image, and then let the model ``correct'' the image. We can similarly draw a sketch and let the model ``fill in'' the sketch with similar details from the input image (see Fig.~\ref{fig:SingleImageTasks}b).
The model is applied to the edited image/sketch by adding noise to the image, and then denoising the input image until a coherent image is obtained.

\vspace*{-0.2cm}
\section{Evaluations \& Comparisons}
\label{sec:evaluations}

This section presents quantitative evaluations to support our main claim for the motion generalization capabilities of SinFusion.
 We measure the performance of our framework by training a model on a small portion of the original video, and test it on unseen frames from a different portion of the same video (Sec.~\ref{sec:next_frame_prediction}). 
 We further propose new useful evaluation metrics for diverse video generation from a single video (Sec.~\ref{sec:diversityMetrics}), and 
compare our diverse video generation from a single video to other methods for this task.

\vspace*{-0.1cm}
\subsection{Future-Frame Prediction from a Single Video}
\label{sec:next_frame_prediction}
Given a video with $N$ frames, we train a model on $n<N$ frames. At inference, we sample $100$ frames from the rest of the $N-n$ frames (not seen during training), and for each of them, use the trained model to predict its next  (or a more distant) frame. We use PSNR to compare a predicted and real frame, and use the average PSNR as the overall score. 

\textbf{Baseline.}
Since no other methods exist for frame-prediction from a single video, we use a simple but strong baseline: Given a frame $f(i)$, we predict its next frame to be identical, namely, $f(i+1)=f(i)$.
This is a strong baseline, since most videos have large static backgrounds, hence there is little change between consecutive frames.

\begin{figure}
    \centering
    \begin{tabular}{c}
        \includegraphics[width=\columnwidth]{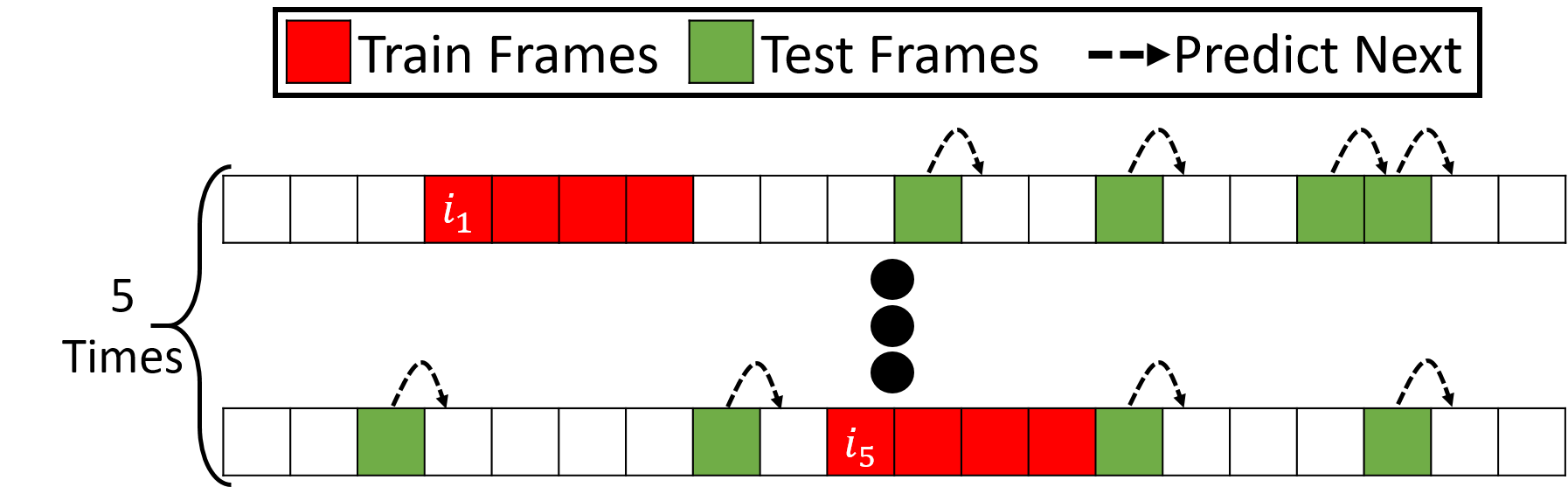} \\
        (a) For each configuration we sample $5$ runs \\ (with different initial frame) \\
        \includegraphics[width=\columnwidth]{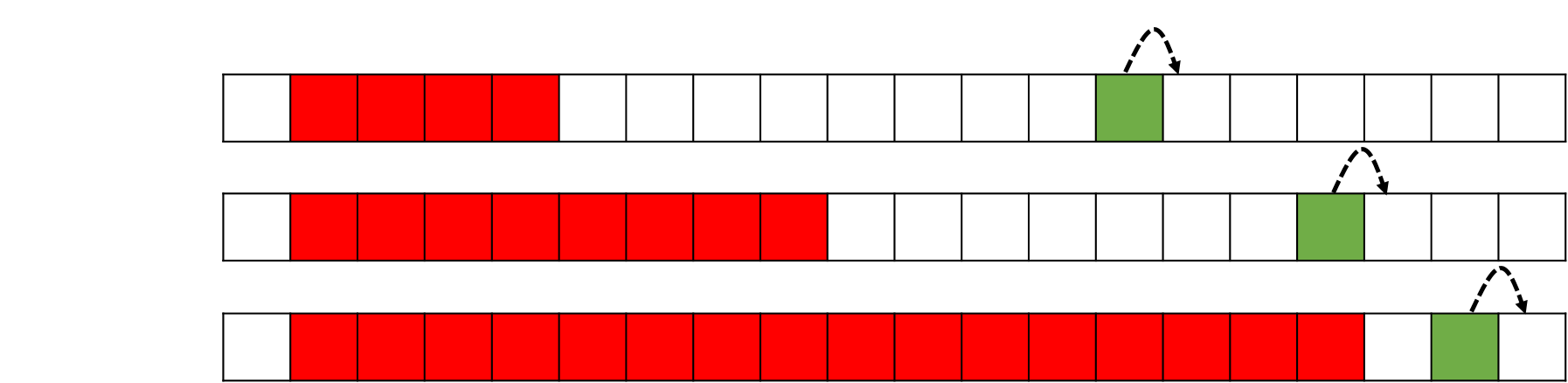} \\
        (b) Different number of training frames \\
        \includegraphics[width=\columnwidth]{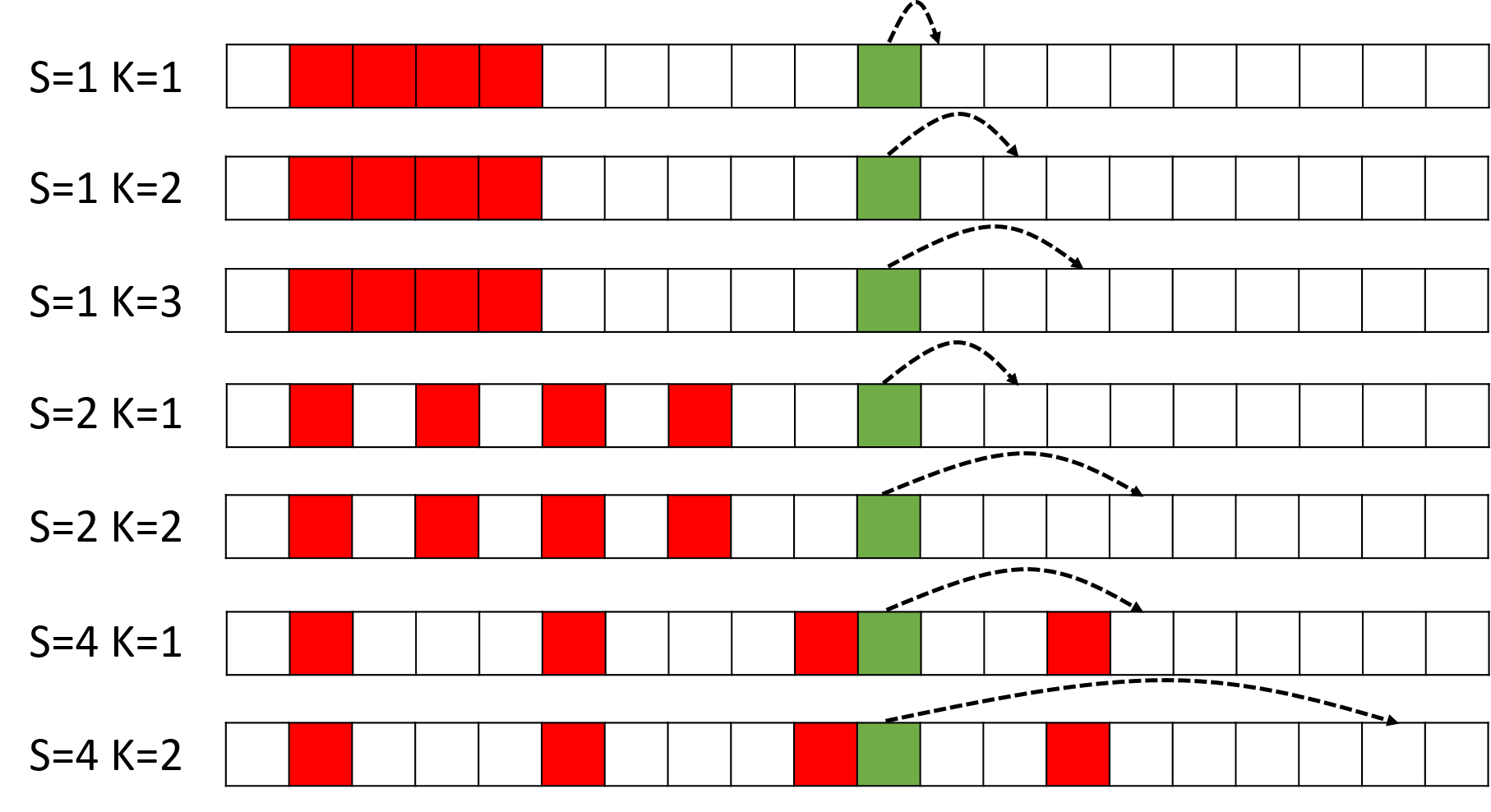} \\
        (c) Different choices of video-speed (S) and $k$
    \end{tabular}
    \vspace*{-0.3cm}
    \caption{\emph{\bf Frame Prediction from a Single-Video}. Depicting evaluations experiments from \cref{sec:next_frame_prediction} and \cref{fig:eval_generalization}}
    \label{fig:evaluations}
    \vspace*{-0.3cm}
\end{figure}

\begin{figure}
    \centering
    \begin{tabular}{c}
    \includegraphics[width=.9\columnwidth]{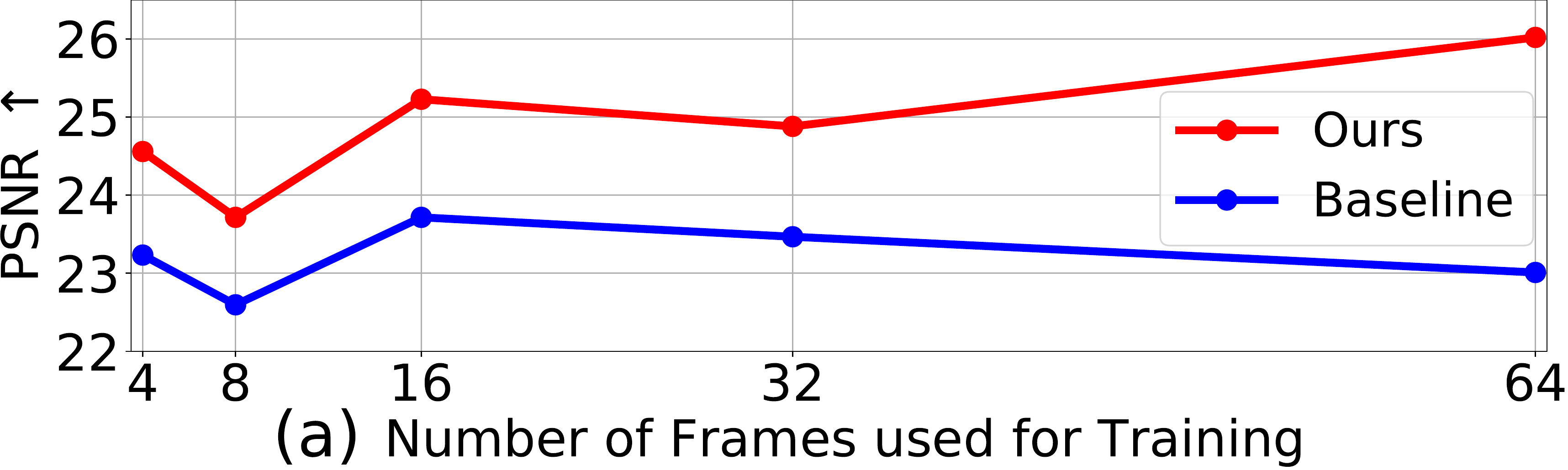} \\
    \includegraphics[width=.9\columnwidth]{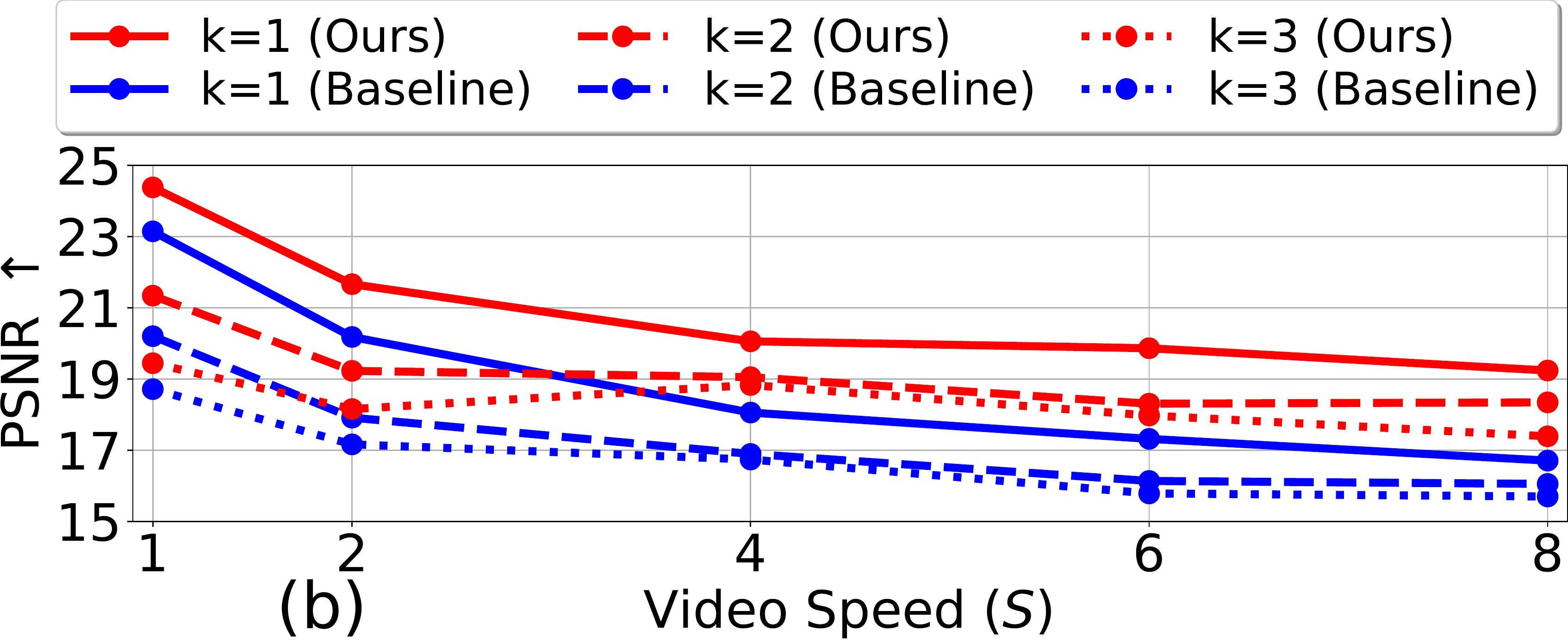}
    \end{tabular}
    \vspace*{-0.3cm}
    \caption{\textbf{Next-Frame Prediction from a Single Video.} SinFusion consistently beats the baseline on this task (see  Sec.~\ref{sec:next_frame_prediction}).
    }
    \label{fig:eval_generalization}
    \centering
    \vspace*{-0.3cm}
\end{figure}

\textbf{Evaluating w.r.t. Different Training Set Sizes (Fig.~\ref{fig:eval_generalization}a):}
We repeat this experiment for varying number of training frames $n$ ($n=[4,8,16,32,64]$).
For each choice of $n$, we choose a random location in the video, and take the $n$ frames starting at that random location
to be the ``training frames''. This is depicted in Fig.~\ref{fig:evaluations}a -- training frames (red) and test frames (green), where each test-frame is used to predict its next frame. In Fig.~\ref{fig:evaluations}b we depict runs trained with different number of training frames $n$. 
The results are shown in Fig.~\ref{fig:eval_generalization}a where each dot corresponds to averaging the score of  $5$ different runs (each time selecting the $n$ training frames at  a different random video location). 
As seen from Fig.~\ref{fig:eval_generalization}a, our framework (red) is consistently better than the baseline (blue), exhibiting the motion prediction/generalization  capabilities of SinFusion.
Note that generalization increases (higher PSNR) with the size of the training set $n$, while the naive baseline does not improve.
Note also that our framework generalizes quite well to next-frame prediction with as few as $n=4$ frames in the training set.

\mbox{\textbf{Evaluating w.r.t. Video ``speed" \& Frame-gap $k$ (Fig.~\ref{fig:eval_generalization}b):}}
We evaluate
how well our framework generalizes on videos with faster motions. To this end, we sub-sample the original video in intervals of increasing size, resulting in faster motions in the sub-sampled videos.
This way we can synthesize videos with larger speeds from the same video, making the results consistent with the first experiment (Fig.~\ref{fig:eval_generalization}a). In this experiment we uses a fixed $n=32$. 

A video with ``speed" $S$ is defined as the original video subsampled at $1/S$. After subsampling the video, the rest of the experiment is carried out as described above. 
For example, if the starting frame is frame number $17$, then the training frames will be frames number $17,19,21,...,79$ for $S=2$, and $17,21,25,...,141$ for $S=4$.

We further evaluate w.r.t. $k$, which is the frame-gap between the current frame and the predicted frame (in the subsampled video) as in Fig.~\ref{fig:video_training}a. Recall that our model trains on $k=[-3,3]$. Several setups for $S$ and $k$ are depicted in Fig.~\ref{fig:evaluations}c.

Results are shown in Fig.~\ref{fig:eval_generalization}b (note that for $S$$=$$1$,$k$$=$$1$, the result is the same as in Fig.~\ref{fig:eval_generalization}a for $n$$=$$32$).
Our framework is consistently better than the baseline. Larger speeds increase the performance gap between our framework and the baseline, further validating our claim for motion generalization.

\subsection{A New Diversity Metric for Single-Video Methods}
\label{sec:diversityMetrics}

\setlength{\tabcolsep}{3pt} 
\begin{table}
    \centering
    \caption{\textbf{Diverse Video Generation -- Comparison.}}
    \vspace*{0.1cm}
    \resizebox{\columnwidth}{!}{
    \begin{tabular}{c|c|ccc}
        Dataset & Method & NNFDIV $\uparrow$ & NNFDIST $\downarrow$ & SVFID$\downarrow$ \\
        \toprule
        \multirow{3}{*}{SinGAN-GIF}  &  VGPNN
        & $0.20$ & $0.28$ & 0.0058\\
        &  SinGAN-GIF
        & $0.40$ & $1.10$ & 0.0119 \\
        &  SinFusion (Ours) & $0.30$ & $0.45$ & 0.0090  \\
        \midrule
        \multirow{3}{*}{HP-VAE-GAN}  &  VGPNN
        & 0.22 & 0.14  & 0.0072 \\
        &  HP-VAE-GAN
        & 0.31 & 0.39 & 0.0081\\
        &  SinFusion (Ours) & 0.35 & 0.26 & 0.0107  \\
    \end{tabular}
    \label{tab:comparison}
    }
    \vspace*{-0.5cm}
\end{table}

We devise a metric to quantify the diversity of our generated samples from a given input video.
SinGAN~\cite{shaham2019singan} proposed the following diversity metric (adapted in a straightforwad manner from images to videos): calculate the standard deviation of the intensity values of each voxel over all generated samples, then average this over all the voxels, and then divide the result by the standard deviation of the intensity values of the original video.

This metric fails on a simple example: given an input video, one could generate ``new" samples by just applying random translations to the video. With enough such ``samples" this will converge to a high diversity score of $1$. Rewarding for such global translations (or ``copies" of large chunks of the input video) is an unwanted artifact of this metric. We introduce a nearest-neighbor-field (\textbf{NNF}) based diversity measure that captures the diversity of generated samples while penalizing for such unnecessary global translations.

\begin{figure}
    \centering
    \includegraphics[width=.9\columnwidth]{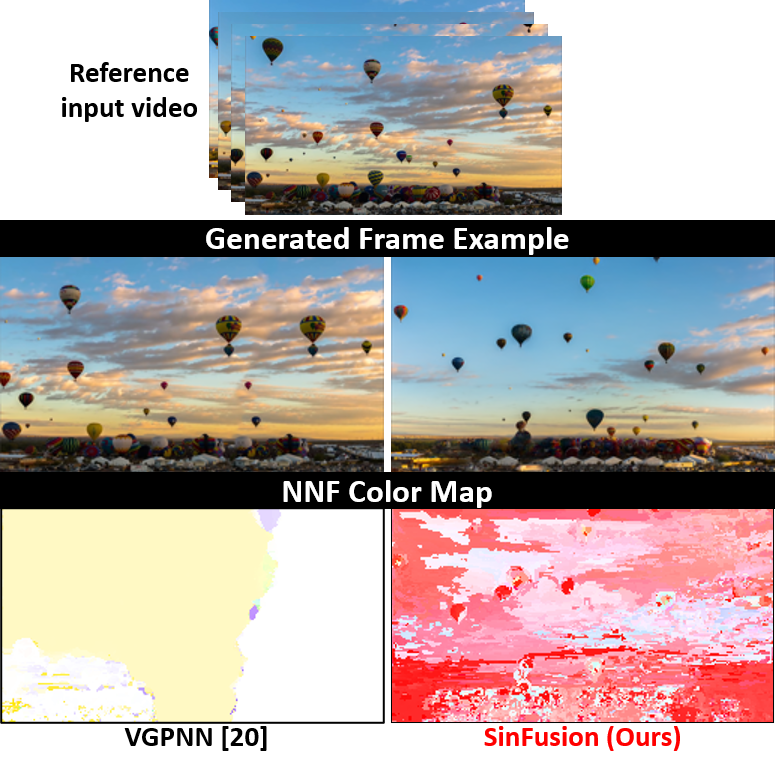}
\vspace*{-0.5cm}
    \caption{\textbf{Nearest-Neighbour Field (NNF) Color Map.}
   Patch-NNF between the generated video and the input video shows that VGPNN~\cite{haim2021diverse} tends to copy large chunks of the input video, whereas SinFusion generates new spatio-temporal compositions.
    }
    \label{fig:nnf_plot}
    \centering
\vspace*{-0.3cm}
\end{figure}

The NNF is computed by searching for each $(3,3,3)$ spatio-temporal patch in a generated video, its nearest-neighbour (\textbf{n.n}) patch in the original video (with MSE). Each voxel is then associated with vector pointing to its n.n. Simple generated videos (e.g. a simple translation of the input) will have a rather constant NNF, while more complex generated videos will have complex NNFs. A visualized example for such NNF is shown in Fig.\ref{fig:nnf_plot} (a vector is converted to RGB using a color wheel~\cite{baker2011database}). See how the NNF of a VGPNN output is simple (corresponds to copying large chunks from a single input frame) whereas ours is more complex (\emph{see full videos of these in our project page}).

We quantify the ``complexity" of an NNF as follows: we use ZLIB~~\cite{gailly2004zlib} to compress the NNF, and record the compression ratio. This gives a diversity measure in $[0,1]$ that we term \emph{\textbf{NNFDIV}}. (The inspiration comes from \emph{Kolmogorov complexity}~\cite{kolmogorov1963tables} -- simpler objects have simpler ``description'', which can be easily bounded by any compression algorithm). We also measure the RGB-similarity (termed \emph{\textbf{NNFDIST}}) by averaging the MSE distance of all generated patch to their n.n's.

In \cref{tab:comparison} we report the results of these metrics, as well as SVFID~\cite{gur2020hierarchical} score, on $2$ diverse video generation datasets (see details in \cref{sec:video_dataset_comparison_appendix}). 
We compare our diverse video generation samples to existing single-video methods -- HP-VAE-GAN~\cite{gur2020hierarchical}, SinGAN-GIF~\cite{arora2021singan} and \mbox{VGPNN}~\cite{haim2021diverse}.

VGPNN is expected to have better quality (low NNFDIST / SVFID) because it is \emph{copying chunks of frames from the original video}. However, its diversity (NNFDIV) is very low. On HP-VAE-GAN dataset, we outperform HP-VAE-GAN in both quality and diversity. On SinGAN-GIF dataset, SinGAN-GIF has higher diversity, however this may be attributed to its very low quality (NNFDIST). For both datasets, SinFusion has the best trade-off in terms of diversity and quality.
\textbf{Further Experiments and Ablations} can be found in Appendices.~\ref{sec:comparison_appendix} and~\ref{sec:ablations_appendix} (e.g., comparison to VDM~\cite{ho2022video}).

\vspace*{-0.3cm}
\section{Limitations}
\label{sec:limitations}
\vspace*{-0.2cm}
As in all single-video generation methods, our method is also limited to videos with relatively small camera motion. Moreover, in videos with large objects of highly non-rigid motions (e.g., with many moving parts), SinFusion may break the object (or remove parts of it). This is because SinFusion has no notion of semantics.
Some of the these limitations may be mitigated by incorporating suitable priors, and is part of our future work.

\vspace*{-0.3cm}
\section{Conclusions}
\vspace*{-0.2cm}
We propose SinFusion, a diffusion-based framework trained on a single video or image. Our unified framework can be applied for a variety of tasks. Our main application -- generation and extrapolation of an input video, exhibits unprecedented generalization capabilities, that were not shown either by previous single-video methods, nor by large-scale video diffusion models.

\section*{Acknowledgements}
We thank Assaf Shocher and Barak Zackay for useful discussions. This project received funding from the European Research Council (ERC) under the European Union’s Horizon 2020 research and innovation programme (grant agreement No 788535) and from the D. Dan and Betty Kahn Foundation.

\bibliography{main}
\bibliographystyle{icml2023}

\clearpage
\newpage

\setcounter{table}{0}
\renewcommand{\thetable}{A\arabic{table}}
\setcounter{figure}{0}
\renewcommand{\thefigure}{A\arabic{figure}}

\begin{figure*}[!ht]
    \captionsetup[sub]{labelformat=parens}
    \captionsetup{type=figure}
    \centering
    \begingroup
        \begin{subfigure}[b]{.49\linewidth}
            \centering
            \includegraphics[width=\textwidth]{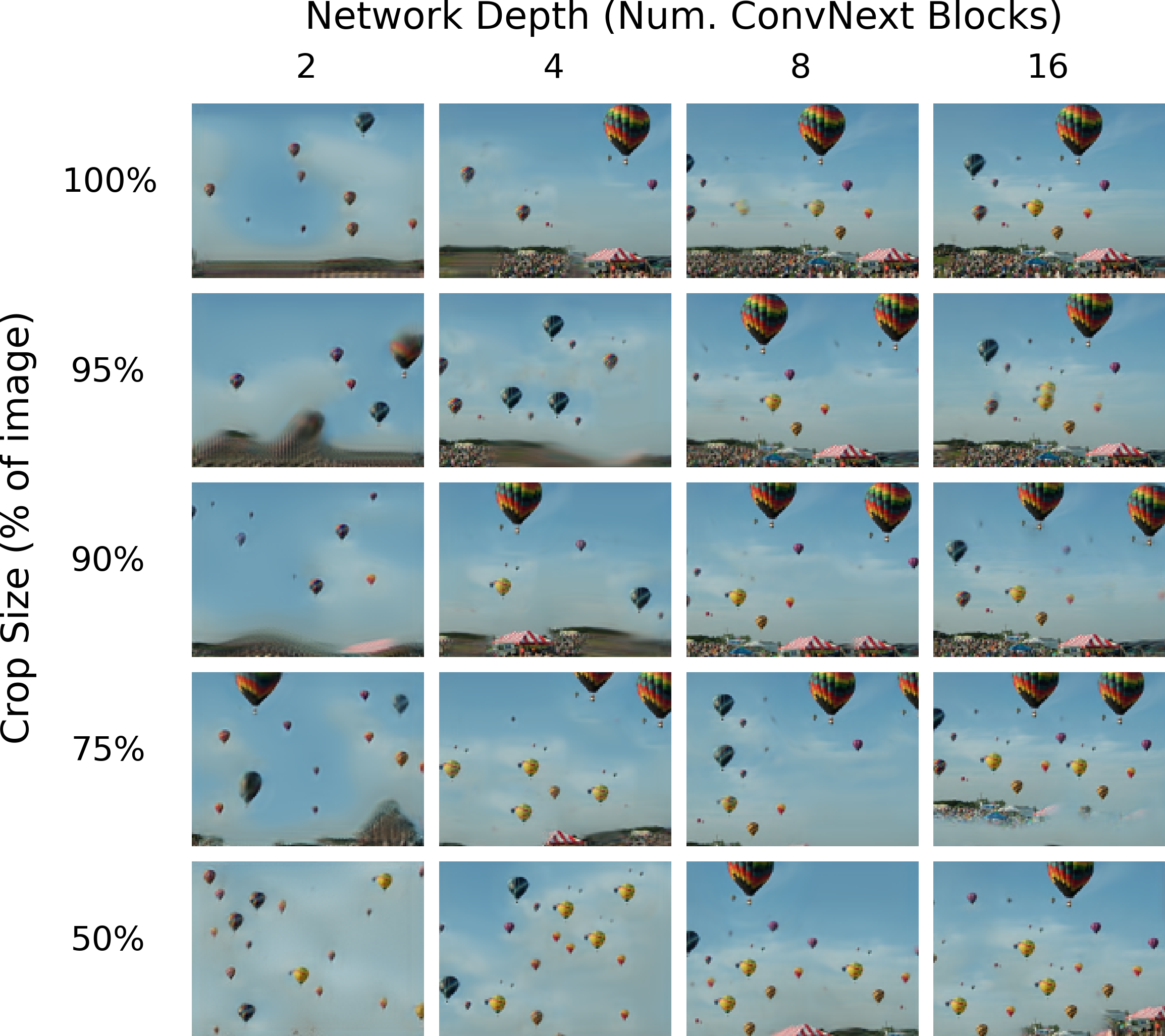}
            \caption{Quality}
        \end{subfigure}
    \endgroup
    \hfill
    \begingroup
        \begin{subfigure}{.49\linewidth}
            \centering
            \raisebox{35pt}{\includegraphics[width=.5\textwidth]{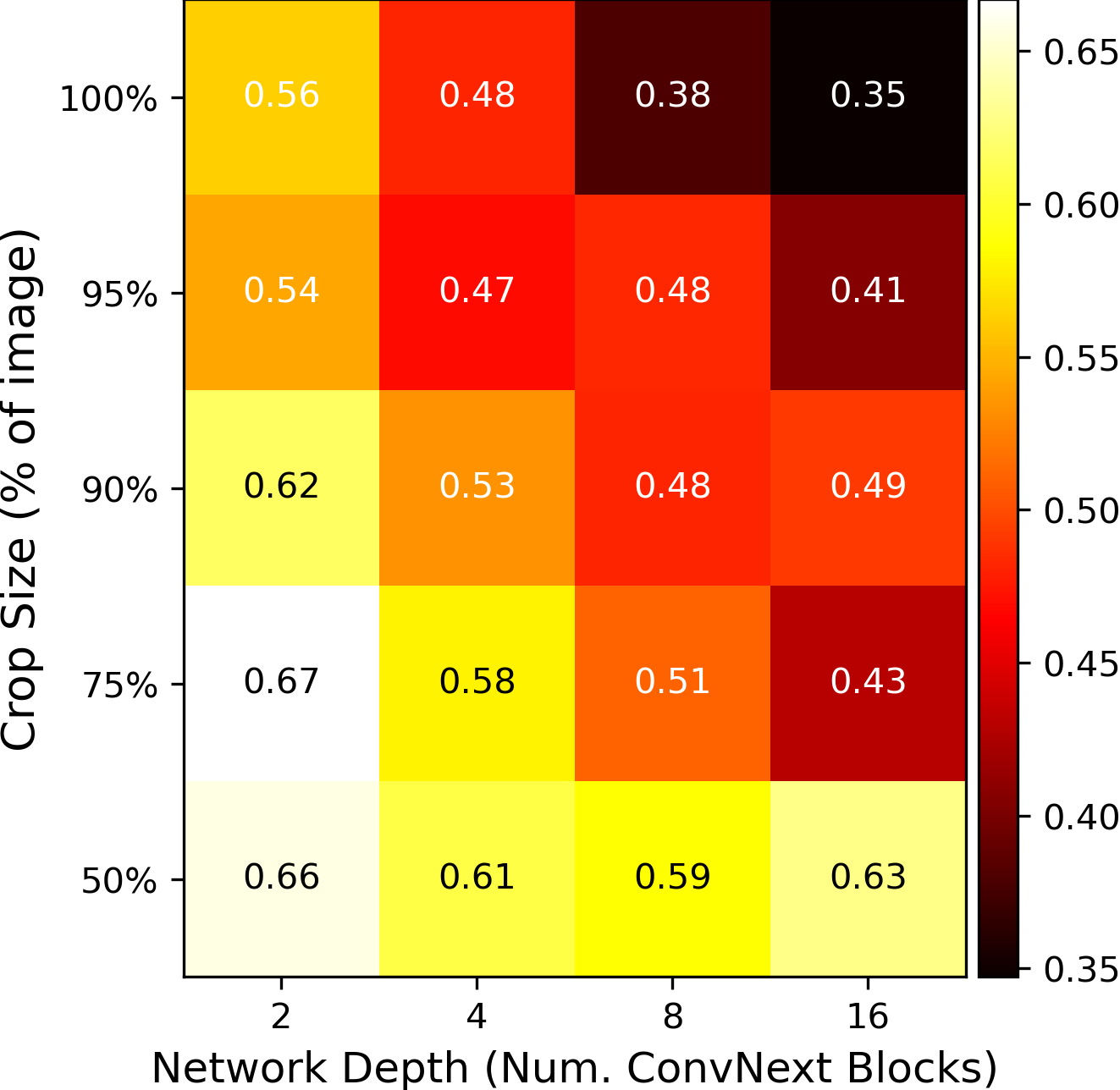}}
            \begin{minipage}{1cm}
            \vfill
            \end{minipage}
            \caption{NNF Diversity (See ~\cref{sec:evaluations})}
        \end{subfigure}
    \endgroup
    \captionof{figure}{Analysing the effect of Crop-Size and Network-Depth on the diversity and quality of the generated outputs}
    \label{fig:diveristy_analysis}
\end{figure*}

\appendix

\addcontentsline{toc}{section}{Appendix} 
\part{Appendix} 
\parttoc 

\section{Further Evaluation - Effect of Crop-Size and Receptive Field}
\label{sec:effect_of_cs_and_rf}

Our goal is to generate outputs (image / video) that preserve global structure, are of high quality, and with large diversity.
These are affected by the choice of the crop-size on which the model is trained, and the effective receptive field of the model (determined by the depth of the convolutional model and controlled via the number of ConvNext blocks in the network). 
As seen in \cref{fig:diveristy_analysis}b, the largest diversity is achieved for small crop-size and small receptive field. However, small networks fail to learn the underlying image structure and result in blurry outputs (\cref{fig:diveristy_analysis}a). We therefore use more blocks for the model. This reduces the diversity, but dramatically improves outputs quality (as is evident from \cref{fig:diveristy_analysis}a). We choose the crop-size as a trade-off to preserve global-structure but also high diversity, which means using crop-size of about $95\%$ of the image, with network depth of $16$ blocks.

\section{Comparisons}
\label{sec:comparison_appendix}

\subsection{Generation from Single-Video (\cref{tab:comparison})}
\label{sec:video_dataset_comparison_appendix}
We run our comparisons on the data provided by the previous works on video generation from a single video: VGPNN~\cite{haim2021diverse}, HP-VAE-GAN~\cite{gur2020hierarchical} and SinGAN-GIF~\cite{arora2021singan}. We follow the same methodology used in VGPNN~\cite{haim2021diverse}.

We compare to two datasets of videos. One provided by SinGAN-GIF ~\cite{arora2021singan} and the other by HP-VAE-GAN~\cite{gur2020hierarchical}. In SinGAN-GIF there are $5$ videos with $8$ to $15$ frames, each of maximal spatial resolution $168\times298$. For each of the $5$ input videos, each method generates $6$ samples.
In HP-VAE-GAN there are $10$ videos each of spatial resolution $144\times256$. For each of the $10$ input videos, each method generates $10$ samples. HP-VAE-GAN and VGPNN only use the first $13$ frames since their methods are limited by runtime and memory. Since learning on small amounts of data is not a goal for the task of diverse generation from a single video, and since our framework can easily learn from much more data, we train our framework on longer sequences of frames from the given input videos.

\subsection{Comparison to VDM~\cite{ho2022video}}
In the project page we show the results of VDM~\cite{ho2022video} trained on a single video. Since the official implementation was not published at the time of writing, we use a third-party implementation \footnote{\url{https://github.com/lucidrains/video-diffusion-pytorch}}). Since VDM expects a dataset of videos, we slice a long video of $420$ frames into $42$ short videos of $10$ frames each and let VDM train on those videos. We could only train the model on a resolution of $64\times64$ pixels before exceeding the memory of our GPU. We trained the model for about $150$ epochs using two different learning rates (total run time was about a day on a V100 GPU).
The results seems to capture the motions of the original videos, but it is difficult to evaluate since the resolution is too low compared to the original videos. The results also contain artifacts that may result from the low amount of data usually needed to train such models (without including our proposed modifications). It is qualitatively evident from the results that our framework, SinFusion, generates outputs of much higher quality when trained on a single video.
For the full video results please see our project page.

\subsection{Comparison to Single Image GANs}
While the main focus our work is on single-video generation/manipulation tasks, we also measure the performance of our single-image DDPM on diverse image generation, in comparison to existing single-image GAN works \cite{shaham2019singan,hinz2021improved}. 
Such a quantitative comparison is presented in \cref{tab:single_image_comparison}. We use the established Single Image FID (SIFID) metric, as well as our NNFDIV metric. We perform the comparison on the Places50 benchmark dataset.
The quantitative comparison shows that our single-image DDPM achieves good performance compared to existing single image generation methods (better in one measure; worse in the other measure). While the Single Image FID achieved by our approach is slightly worse than the competing methods, we attribute that to the fact that our model generalizes beyond the internal patch distribution of the single training image (as evident in our better diversity score NNFDIV).

An additional advantage of our single-image DDPM, when compared to single-image GANs, stems from the boundary bias that exists in SinGAN and ConSinGAN \cite{xu2020positional}. This induced bias causes fixed content in corner regions of the generated images, which hurts diversity. This boundary bias occurs because the discriminator (in each scale) of single-image GANs sees the same ground-truth image in all training iterations. Thus, the generator "learns" to output the same boundary as in the original image, by relying on the padding of the training image. In contrast, our single image DDPM trains on different crops of the input image, hence it does not suffer from this bias, and produces more diverse outputs, as is seen the higher diversity score in \cref{tab:single_image_comparison}.

\setlength{\tabcolsep}{3pt} 
\begin{table}
    \centering
    \vspace*{0.1cm}
    \begin{tabular}{l|cc}
        Method & SIFID$\downarrow$ & NNFDIV $\uparrow$ \\
        \toprule
        SinGAN & 0.085 & 0.280 \\
        ConSinGAN & 0.072 & 0.315\\
        SinFusion (Ours) & 0.110 & 0.341\\
    \end{tabular}
    \caption{\textbf{Diverse Image Generation -- Comparison.}}
    \label{tab:single_image_comparison}
    \vspace*{-0.5cm}
\end{table}

\section{Ablations}
\label{sec:ablations_appendix}
\begin{figure*}
    \centering
    \begin{tabular}{c|cccc}
        \ccred \includegraphics[width=.19\textwidth]{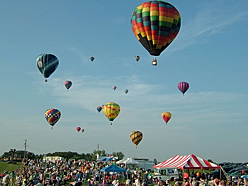}
        &
        \ccpurple \includegraphics[width=.19\textwidth]{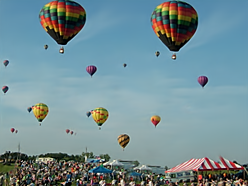}
        &
        \ccpurple \includegraphics[width=.19\textwidth]{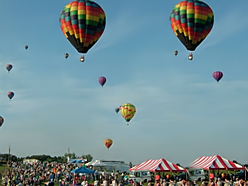}
        &
        \ccpurple \includegraphics[width=.19\textwidth]{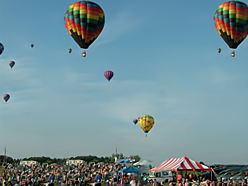}
        &
        \ccpurple \includegraphics[width=.19\textwidth]{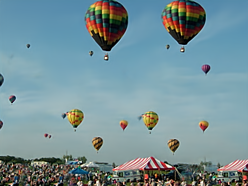}
        \\
        &
        \includegraphics[width=.19\textwidth]{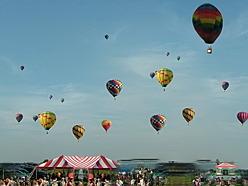}
        &
        \includegraphics[width=.19\textwidth]{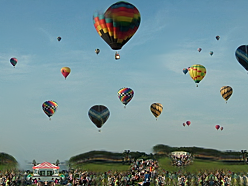}
        &
        \includegraphics[width=.19\textwidth]{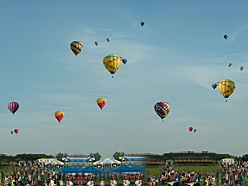}
        &
        \includegraphics[width=.19\textwidth]{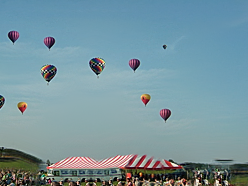}
    \end{tabular}
    \caption{\textbf{Noise vs Image prediction ablation}. \emph{Top Row}: Input image (left;red) and generated outputs of our final model (right;purple) -- predicts the un-noised image crop. \emph{Bottom Row}: Generated outputs using the standard DDPM noise prediction.}
    \label{fig:noise_vs_image}
\end{figure*}

\paragraph{Predicting Image Instead of Noise.}
As opposed to the standard DDPM~\cite{ho2020denoising} training, our single-image-DDPM model outputs a prediction for the un-noised input crop. In \cref{fig:noise_vs_image} we show examples for our generated outputs for predicting the crop/image (top) against generated outputs for predicting noise (bottom). As shown, predicting the un-noised crop instead of the noise generates higher quality images. This is also evident in \cref{tab:quantitative_ablations}, where noise prediction leads to a worse SIFID score. 
In addition, predicting the image instead of noise also converges with much fewer training iterations, a feat we attribute to the lower complexity of the patch distribution of the training image compared to the patch distribution of random noise.

\begin{figure*}
    \centering
    \begin{tabular}{c|cccc}
        \ccred \includegraphics[width=.19\textwidth]{figs/balloons.png}
        &
        \ccpurple \includegraphics[width=.19\textwidth]{figs/image_best/0_sample.png}
        &
        \ccpurple \includegraphics[width=.19\textwidth]{figs/image_best/1_sample.png}
        &
        \ccpurple \includegraphics[width=.19\textwidth]{figs/image_best/2_sample.png}
        &
        \ccpurple \includegraphics[width=.19\textwidth]{figs/image_best/3_sample.png}
        \\
        &
        \includegraphics[width=.19\textwidth]{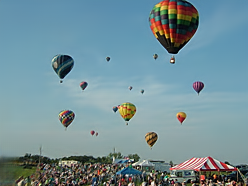}
        &
        \includegraphics[width=.19\textwidth]{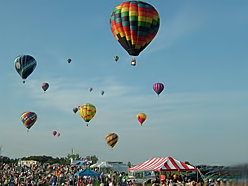}
        &
        \includegraphics[width=.19\textwidth]{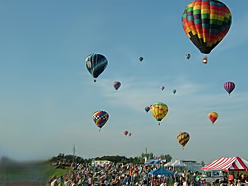}
        &
        \includegraphics[width=.19\textwidth]{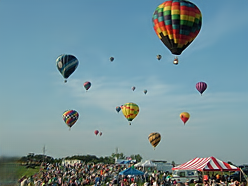}
        \\
        &
        \includegraphics[width=.19\textwidth]{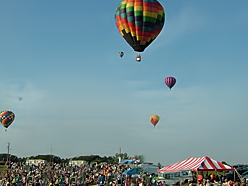}
        &
        \includegraphics[width=.19\textwidth]{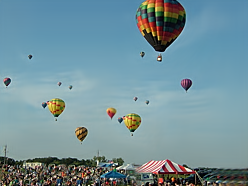}
        &
        \includegraphics[width=.19\textwidth]{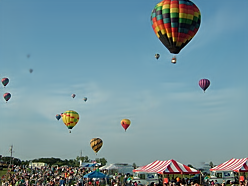}
        &
        \includegraphics[width=.19\textwidth]{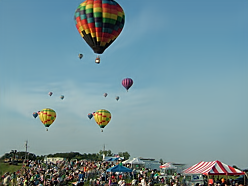}
        \\
        
        &
        \includegraphics[width=.19\textwidth]{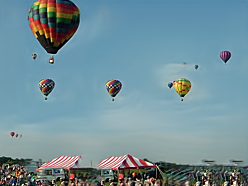}
        &
        \includegraphics[width=.19\textwidth]{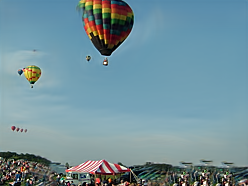}
        &
        \includegraphics[width=.19\textwidth]{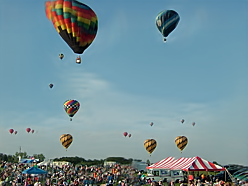}
        &
        \includegraphics[width=.19\textwidth]{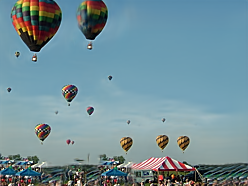}
        \\
    \end{tabular}
    \caption{\textbf{Architectural ablations}. 
    \emph{Top Row}: Input image (left;red) and Generated outputs of our final model (right;purple).\newline
    \emph{2nd Row}: Generated outputs of our model with upsampling and downsampling layers. \newline
    \emph{3rd Row}: Generated outputs of our model with added attention layers (similar to the standard DDPM \cite{ho2020denoising} Unet\cite{ronneberger2015u} network). \newline
    \emph{4th Row}: Generated outputs of our model where each ConvNext \cite{liu2022convnet} block is replaced with a ResNet \cite{he2016deep} block.
    }
    \label{fig:arch_ablations}
\end{figure*}

\paragraph{Architectural changes.}
We check the importance of our proposed architectural modifications to the original DDPM \cite{ho2020denoising} by reverting each change and generating several images. We compare these generated images to the images generated by our final model. An example for these comparisons can be seen in \cref{fig:arch_ablations}. The quantitative results for these ablations can be found in \cref{tab:quantitative_ablations}. \\
The first comparison shows outputs of our model with upsampling and downsampling layers. The generated outputs completely overfit the training image, and have no diversity. This is also evident in the low NNFDIV score in \cref{tab:quantitative_ablations}.\\
The second comparison shows outputs of our model with attention layers. Other than significantly increasing training time (to almost 2 hours per training image), the added attention decreases the quality and diversity of the generated samples, as evident in \cref{tab:quantitative_ablations}.\\
The third comparison shows outputs of our model with all ConvNext \cite{liu2022convnet} blocks replaced with ResNet \cite{he2016deep} blocks. The generated outputs suffer from smearing artifacts and are of lesser quality than our generated outputs, as also evident by the lower SIFID in \cref{tab:quantitative_ablations}.

\begin{table}
    \centering
    \vspace*{0.1cm}
    \begin{tabular}{l|cc}
        Model & SIFID$\downarrow$ & NNFDIV $\uparrow$ \\
        \toprule
        Our single-image DDPM & 0.181 & 0.480\\
        Noise Prediction & 0.473 & 0.506 \\
        w/ Up/Down sampling layers & 0.145 & 0.258 \\
        w/ Attention layers & 0.256 & 0.396 \\
        w/ ResNet blocks & 0.246 & 0.463 \\
    \end{tabular}
    \caption{\textbf{Architectural changes and noise prediction ablations.} We ablate our design choices by measuring the quality (via SIFID) and diversity (via NNFDIV) generated images. The results show that our final single-image DDPM achieves the best tradeoff between generation quality and the diversity of the generated samples.}
    \label{tab:quantitative_ablations}
    \vspace*{-0.5cm}
\end{table}

\paragraph{Importance of DDPM frame Projector in diverse video generation.}
We show the necessity of our DDPM frame Projector model as part of the diverse video generation framework. In this ablation, we generate videos from several input videos using only the DDPM frame Predictor to generate frames, without using the Projector model to correct small artifacts in the generated frames. In all examples, it can be seen that the small artifacts, which remain uncorrected, accumulate over time and severely degrade the generation quality. The quantitative results can be seen in \cref{tab:projector_ablation}, where we measure the SVFID score of the generated videos. For qualitative video results please see the project page.

\begin{table}
    \centering
    \begin{tabular}{l|cc}
        \multicolumn{1}{c|}{Dataset} & \multicolumn{2}{c}{SVFID$\downarrow$} \\
        & With Projector & No Projector  \\
        \toprule
        All videos in project page & \bf{0.0066} & 0.0081\\
        HP-VAE-GAN & \bf{0.0107} & 0.0129\\
        SinGAN-GIF & \bf{0.0090} & 0.0136 \\
    \end{tabular}
    \caption{\textbf{DDPM frame Projector ablation.} The DDPM frame Projector consistently improves the quality of the generated videos, as evident by the lower SVFID scores.}
    \label{tab:projector_ablation}
    \vspace*{-0.5cm}
\end{table}

\paragraph{Effect of training with $k=[-3,3]$.} As written in~\cref{sec:single_video_ddpm}, 
at inference time we always use either $k=1$ (for forward prediction) or $k=-1$ (for backward prediction). However, we found that training the predictor with $k\in[-3,3]$ improves the prediction for $k=\pm 1$. For example, training the predictor with only results in SVFID = 0.0112 (averaged on all videos in the supplementary), whereas training it with results in SVFID = 0.0095 (lower SVFID is better).

\section{Further Explanations on Related Works}

In this section we elaborate further on existing methods.

\paragraph{Diffusion models for Videos:}
\label{sec:diffusion_videos}
\begin{itemize}
    \item RVD~\cite{yang2022diffusion} tackles video prediction by conditioning the generative process on recurrent neural networks.
    \item RaMViD~\cite{hoppe2022diffusion} and MCVD~\cite{voleti2022mcvd} train an autoregressive model conditioned on previous frames for video prediction and infilling using masking mechanisms.
    \item VDM~\cite{ho2022video} introduces unconditional video generation by modifying the Conv2D layers in the basic DDPM UNet to Conv3D, as well as autoregressive generation.
    \item Imagen-Video~\cite{ho2022imagen} extends VDM to text-to-video and also include spatio-temporal superresolution conditioned on upsampled versions of smaller scales. 
    \item FDM~\cite{harvey2022flexible} modifies DDPM to include temporal attention mechanism and can be conditioned on any number of previous frames.
\end{itemize}

\paragraph{Generation from a Single Image.}
Generative models trained on a single image aim to generate new diverse samples, similar in appearance to the image/video on which they were trained. 
Most notably, SinGAN~\cite{shaham2019singan} and InGAN~\cite{shocher2018ingan} trained multi-scale GANs to learn the distribution of patches in an image. They showed its applicability to diverse random generation from a single image, as well as a variety of other image synthesis applications (inpainting, style transfer, etc.). Their results are usually better suited to synthesis from a single image than models trained on large collection of data.
More recently, GPNN~\cite{granot2021drop} showed that most image synthesis tasks proposed by single-image GAN-based models~\cite{hinz2021improved,shaham2019singan,shocher2018ingan} can be solved by classical non-parametric patch nearest-neighbour methods~\cite{efros1999texture,efros2001image,simakov2008summarizing}, and achieve outputs of higher quality while reducing generation time by orders of magnitude. 
However, nearest-neighbour methods have a very limited notion of generalization, and are therefore limited to tasks where it is natural to "copy" parts of the input.
In this respect, learning based methods like SinGAN~\cite{shaham2019singan} 
still offer applicability like shown in the tasks of harmonization or animation. 

\vspace*{-0.3cm}
\paragraph{Generation from a Single Video.}
Similar to the image domain, extensions of SinGAN~\cite{shaham2019singan} to generation from a single \emph{video} were proposed~\cite{gur2020hierarchical,arora2021singan}, generating diverse new videos of similar appearance and dynamics to the input video. These too, were outperformed by patch nearest-neighbour methods~\cite{haim2021diverse} in both output quality and speed.
However, these video-based nearest-neighbour methods suffer from drawbacks similar to the image case. While the generated samples are of high quality and look realistic, the main reason for this is that the samples are essentially copies of parts of the original video stitched together. They fail to exhibit motion generalization capabilities. 
None of the above-mentioned methods can handle input videos longer than a few dozens frames. Single-video GAN based methods are limited in compute time (e.g., HP-VAE-GAN~\cite{gur2020hierarchical} takes 8 days to train on a short video of $13$ frames), whereas VGPNN~\cite{haim2021diverse} is limited in memory (since each space-time patch in the output video searches for its nearest-neighbor space-time patch in the entire input video, at each iteration).
In contrast, our method can handle any length of input video. While it can generalize well from just a few frames, it can also easily train on a long input video at a fixed and very small memory print, and at reasonable compute time (a few hours per video).

\section{Implementation Details}
\label{sec:implemetation_details}
Our code is implemented with PyTorch~\cite{paszke2017automatic}.
We make the following hyper-parameters choices:
\begin{itemize}
    \item We use a batch size of $1$. Each large crop contains many large "patches". Since our network is a fully convolutional network, each large "patch" is a single training example. 
    \item We use ADAM optimizer~\cite{kingma2014adam} with a learning rate of $2 \times 10^{-4}$, reduced to $2 \times 10^{-5}$ after $100K$ iterations. 
    \item We set the diffusion timesteps $T = 50$. This allows for fast sampling, without sacrificing image/video quality (This trade-off is simpler in our case because of the simplicity of our learned data distribution).
    \item When generating diverse videos, we use the DDPM frame Projector to correct predicted frames by noising and denoising $T_{corr}=3$ steps.
    \item We compared several noise schedules for the diffusion models and ended up using linear noise schedule ($\beta_0=2\times10^{-3}, \beta_T=0.4$) for single-image DDPM and cosine noise schedule \cite{nichol2021improved} for single-video DDPM.
    \item Our standard network architecture consists of $16$ ConvNext \cite{liu2022convnet} blocks, each block with a base dimension of $64$ channels.
\end{itemize}

\subsection{Runtimes}
\label{sec:runtimes}
On a Tesla V100-PCIE-16GB, for images/videos of resolution $144 \times 256$, our model trains for about $1.5$ minutes per $1000$ iterations, where each iteration is running one diffusion step on a large image crop. The total amount of iterations and total runtime for each of our models are:
\begin{itemize}
    \item Single-Image DDPM - $50K$ iterations, total runtime of $80$ minutes (good results are already seen after $15K$ iterations).
    \item Single-Video DDPM Frame Predictor - $200K$ iterations, total runtime of $5.5$ hours.
    \item Single-Video DDPM Frame Projector - $100K$ iterations, total runtime of $2.5$ hours
    \item Single-Video DDPM Frame Interpolator - $50K$ iterations, total runtime of $1.5$ hours.
\end{itemize}

\section{Videos Sources}
In our project page we show results for video generation and and extrapolation for videos excerpts from the following YouTube videos (YouTube video IDs): \\
$\bullet$  LkrnpO5v0z8  \\   
$\bullet$  hj6EG7x-BT8  \\  
$\bullet$  nRxSUkZYeOE  \\  
$\bullet$  9ePic3dtykk  \\  
$\bullet$  pB6XSixrCC8  \\  
$\bullet$  ZO5lV0gh5i4  \\  
$\bullet$  tmPqO\_TGa-U \\   
$\bullet$  bsSypB9gI0s  \\  
$\bullet$  RZ1kK-X3QwM  \\  
$\bullet$  FR5l48\_h5Eo \\   
$\bullet$  4i6VSrIYRYY  \\  
$\bullet$  m\_e7jUfvt-I \\   
$\bullet$  DniKM5SKe6c  \\  
$\bullet$  rbzxxbuk3sk  \\  
$\bullet$  W\_yWqFYSggc \\   
$\bullet$  WA5fqO6LUUQ  \\  
$\bullet$  -ydgKb5K\_kc \\   

We also use several videos from MEAD Faces Dataset~\cite{wang2020mead}, and Timlapse Clouds Dataset~\cite{jacobs2010using,jacobs2013two}.

\end{document}